\documentclass[journal]{IEEEtran}

\usepackage{array}
\usepackage{booktabs}
\usepackage{cite}
\usepackage{graphicx}
\usepackage{multirow}

\newcommand\etal{\emph{et~al.}}
\newcommand\ie{\emph{i.e.}}

\newcommand\eg{\emph{e.g.}}

\begin{document}

\title{Fast Video Crowd Counting \\with a Temporal Aware Network}

\author{Xingjiao Wu, Baohan Xu, Yingbin Zheng, Hao Ye, Jing Yang, Liang He
\thanks{Xingjiao Wu and Baohan Xu contributed equally to this work. Corresponding author: Jing Yang (e-mail: jyang@cs.ecnu.edu.cn).}
\thanks{X. Wu, J. Yang, and L. He are with East China Normal University, Shanghai 200062, China.}
\thanks{B. Xu is with Jilian Technology Group (Video++), Shanghai 200023, China.}
\thanks{Y. Zheng and H. Ye are with Videt Tech Ltd., Shanghai 201203, China.}
}

\maketitle

\begin{abstract}
Crowd counting aims to count the number of instantaneous people in a crowded space, and many promising solutions have been proposed for single image crowd counting. With the ubiquitous video capture devices in public safety field, how to effectively apply the crowd counting technique to video content has become an urgent problem. In this paper, we introduce a novel framework based on temporal aware modeling of the relationship between video frames. The proposed network contains a few dilated residual blocks, and each of them consists of the layers that compute the temporal convolutions of features from the adjacent frames to improve the prediction. To alleviate the expensive computation and satisfy the demand of fast video crowd counting, we also introduce a lightweight network to balance the computational cost with representation ability. We conduct experiments on the crowd counting benchmarks and demonstrate its superiority in terms of effectiveness and efficiency over previous video-based approaches.

\end{abstract}


\IEEEpeerreviewmaketitle

\section{Introduction}
\label{sec:intro}

The rapid development of surveillance devices has led to an explosive growth of images and videos, which creates a great demand for analyzing visual content. In addition to object recognition, crowd counting, which focuses on estimating the number of people from the visual contents, has received increasing interests in recent years. Many researchers have explored crowd counting task on still images, while limited efforts have been focused on videos. Nevertheless, crowd counting in videos has many real-world applications, such as video surveillance, traffic monitoring, and emergency management.

Counting crowd robustly and efficiently under different pedestrian distribution, illumination, occlusion, and camera distortion is nevertheless challenging. Although recent progresses such as multi-branch network are introduced to learn more contextual information and achieve excellent performance, most of existing methods still ignore the temporal relations between nearby frame since crowd counting data often collected by surveillance videos. Temporal relation is an important factor in video tasks comparing with still images. There is a certain overlap between video frames, so that there is a certain law between continuous data. Exploring and using temporal relations can correct some errors caused by noise. From the results, our experiments show that considering temporal relations performs significantly better than most current methods which ignore the relationships. Furthermore, the predicted density maps, which may be the most common intermediate element in a modern crowd counting system, is also with a strong correlation with the neighboring maps in the video sequence.

\begin{figure*}
  \centering
  \includegraphics[width=.9\linewidth]{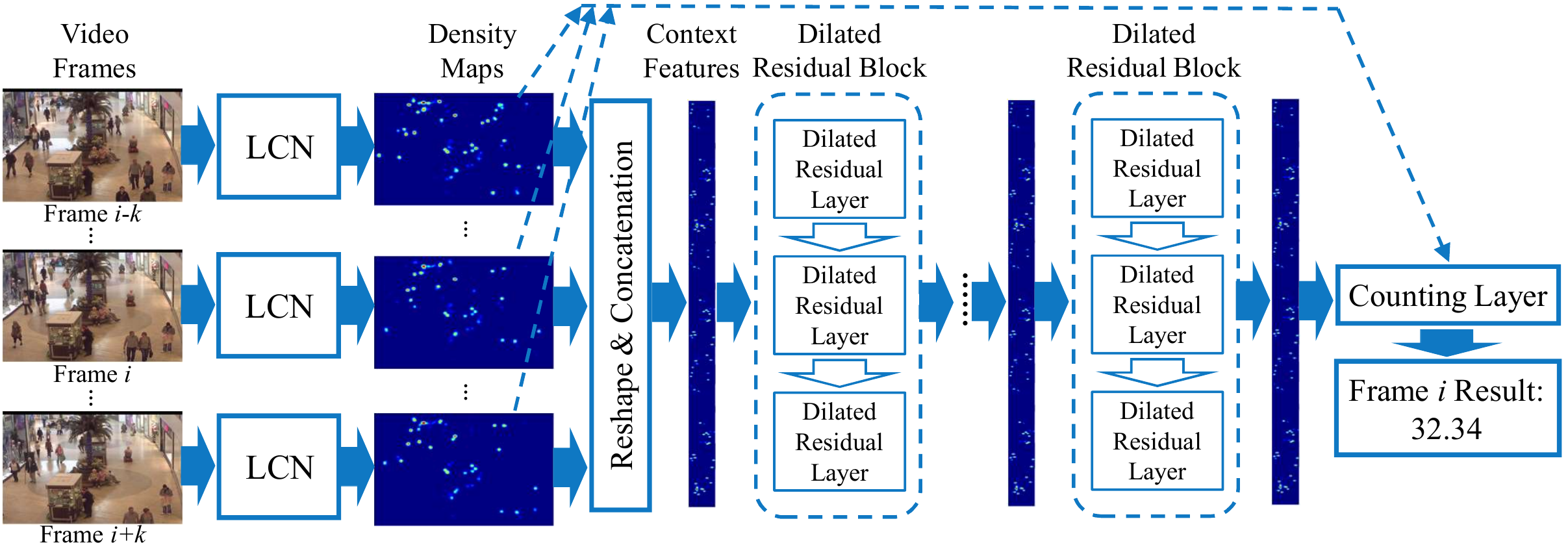}
  \caption{The Architecture of proposed TAN. A lightweight convolution network is used to produce density map for each frame efficiently. Then the dilated residual blocks explore both short-term and long-term context information gain from adjacent density maps as weight vectors. The final output is computed by multiply the frame features with weight vectors.}
  \label{fig:framework}
\end{figure*}

To cope with these difficulties, we employ a novel framework to take advantage of temporal information extracted by continuously video frames. The Temporal Aware Network (TAN) is proposed to dynamically simulate the temporal characteristics of continuous frames for crowd counting and the architecture is shown in Fig. \ref{fig:framework}. The TAN consists of two main parts, an lightweight convolutional neural network (LCN) unit capable of processing counting tasks quickly, and a multiple block architecture for temporal modeling. The LCN unit can guarantee the network response speed while keeping a certain accuracy. Then we focus on modeling the relationship in time dimension and construct a group of dilated residual blocks between the adjacent features. Within each dilated residual block, we employ an expanded set of temporal convolutions to update the frame-level prediction with the help of its neighboring frames. Different from the existing works, we also introduce the density map as another branch of our architecture. Our observation is, while the adjacent video frames may have different visual content due to the background and occlusion, the neighboring density maps still demonstrate more similar content with each other. The density map reports the distribution of people, which can be regarded as attention map. The contextual information between consequent frames and density maps would benefit the current counting state. Comprehensive experiments on the public datasets show the improvement with the help of temporal and contextual information.

The main contributions of this work are summarized as follows.
\begin{itemize}
    \item The proposed Temporal Aware Network dynamically model the temporal features from continuously video frames for crowd counting. Utilizing information from density maps helps to overcome the changing backgrounds and occlusion and boosts the performance.
    \item We also design a lightweight convolutional network to achieve a comparable result while keeping the compactness of the model.
    \item Extensive evaluations on the benchmarks demonstrate the superior performance of our proposed method. Notably, we achieve the state-of-the-art results on the video datasets comparing with the existing video-based methods. Furthermore, our network achieves 25 FPS crowd counting speed on a moderate commercial CPU.
\end{itemize}

The rest of this paper is organized as follows. Section \ref{sec:related} introduces background of crowd counting in images and videos. Section \ref{sec:app} discusses the model design, network architecture and training process in detail. In Section \ref{sec:exp}, we demonstrate the qualitative and quantitative study of the proposed framework. We conclude our work in Section \ref{sec:conclusion}.

\section{Related Work}
\label{sec:related}

\subsection{Crowd Counting in Still Images}

Over the past few years, researchers have attempted to solve crowd counting in images using a variety of approaches. Early works focused on detection methods to recognize specific body parts or full body using hand-crafted features~\cite{dalal2005histograms, li2008estimating}. While detection based methods are difficult to deal with dense crowds because of occlusion, some studies investigated to learn a mapping function between features to the number of peoples~\cite{chan2008privacy}. Furthermore, Lempitsky~\etal~\cite{lempitsky2010learning} proposed local features for the density map to exploit spatial information. However, the hand-crafted features are not good enough when facing the clutter and low resolution of images.

Recently, the convolutional neural network has shown great success in computer vision fields. Inspired by the promising performance of the neural network, many researchers have explored CNN-based methods in crowd counting~(\eg, \cite{Idrees2013Multi,Zhang_2016_CVPR,onoro2016,Sam_2017_CVPR,Sindagi2017Generating,Li2018CSRNet,Shi_2018_CVPR,liu2018leveraging,babu2018top,ranjan2018iterative,idrees2018composition,zou2019attend,gao2019scar,cheng2019learning,xu2019learn,liu2019crowd,wu2019adaptive,ma2019atrous,wang2019removing,zhang2019multi,wang2019learning,gao2019pcc,wang2020detecting}). Zhang~\etal~\cite{Zhang_2016_CVPR} proposed a multi-column CNN with different sizes of filters to deal with the variations of density differences. Sam~\etal~\cite{Sam_2017_CVPR} and Sindagi~\etal~\cite{Sindagi2017Generating} have achieved remarkable results in a multi-subnet structure. To address the problem of limited training data, Liu~\etal~\cite{liu2018leveraging} investigated enhance data such as collect scene datasets from Google using keyword searches and query-by-example image retrieval and then applying a learning-to-rank method. Shi~\etal~\cite{Shi_2018_CVPR} considered that the adaptation of the previous method to the crowd relying on a single image is still in its infancy. Sam~\etal~\cite{babu2018top} proposed the TDF-CNN with top-down feedback to correct the initial prediction of the CNN that is very limited for detecting the space background of people. These methods are all designed for image crowd counting, thus treating videos as image sequences would ignore the important temporal information in videos. Wang \etal~\cite{wang2019learning} proposed a crowd counting method via domain adaptation. The data collector and labeler can generate the synthetic crowd scenes and simultaneously annotate them without any manpower. Gao \etal~\cite{gao2019pcc} proposed the perspective crowd counting network to overcome the deficiency of traditional methods that only focus on the local appearance features. Recently, a novel structural context descriptor was designed to characterize the structural characteristics of individuals in crowd scenes and make better use of context information~\cite{wang2020detecting}.

\subsection{Crowd Counting in Videos}

There are fewer researchers studied on video crowd counting compared with still images. Brostow~\etal~\cite{brostow2006unsupervised} and Chan~\etal~\cite{chan2012counting} proposed to use the Bayesian function to detect individuals using motion information. Rodriguez~\etal~\cite{rodriguez2011density} further proposed optimization of energy function combining crowd density estimation and individual tracking. Chen~\etal~\cite{chen2015person} proposed an error-driven iteration framework aiming to cope with the noisy input videos.

Although these methods based on motion or hand-crafted features showed satisfactory performance on the pedestrian or football datasets, they are still lack of the generalization ability when applying them to extremely dense crowds. More recently, Xiong~\etal~\cite{xiong2017spatiotemporal} proposed the convLSTM framework to capture both spatial and temporal dependencies. The CNN-based method demonstrated the effectiveness of benchmark crowd counting datasets, such as UCF\_CC\_50~\cite{Idrees2013Multi} and UCSD~\cite{chan2008privacy}. However, due to the limited training data of videos and various scenes, it is usually difficult to train the complex and deeper networks for effective crowd counting. In this paper, we propose a novel framework considering temporal information and density maps as well. Even though using a lightweight network architecture, our method can achieve promising results on multiple datasets, with the help of the auxiliary information extracted from temporal dependencies and density maps.

\section{Framework}
\label{sec:app}

In this section, we will introduce the temporal aware modeling with the convolutional network for the crowd counting task in the video. We describe architecture of temporal modeling in Section~\ref{sec:DTM} and the basic unit of the temporal aware network in Section~\ref{sec:LCNN}. The implementation details will be described in Section~\ref{sec:ID}.

\subsection{Temporal Aware Network}
\label{sec:DTM}

The selection of a temporal modeling approach is important to the success of the video crowd counting system. Ideally, we want a comprehensive collection of both long-term and short-term frame correlations so that we can have accurate counting under any scene settings. However, video processing is time-consuming and the training video dataset for crowd counting is also limited. With these in mind, we design the Temporal Aware Network (TAN) with the dilated convolution to fully utilize the context and content information of the video. The architecture of TAN is shown in Fig. \ref{fig:framework}. Different from the existing works, we mainly focus on investigating the relations between combination density maps rather than only considering the temporal information of original frames.

Vectors from several neighboring video frames are concatenated as the inputs of the first dilated block. Particularly, for the $t$-th video frame, we suppose $k$ frames before and after the frame are considered and the input feature for the $t$-th frame is ${\mathbf{v}}_0(t) = [{v_{t - k}^T}, \cdots,v_t^T \cdots,{v_{t + k}^T}]^T$.

\begin{figure}
  \centering
  \includegraphics[width=0.7\linewidth]{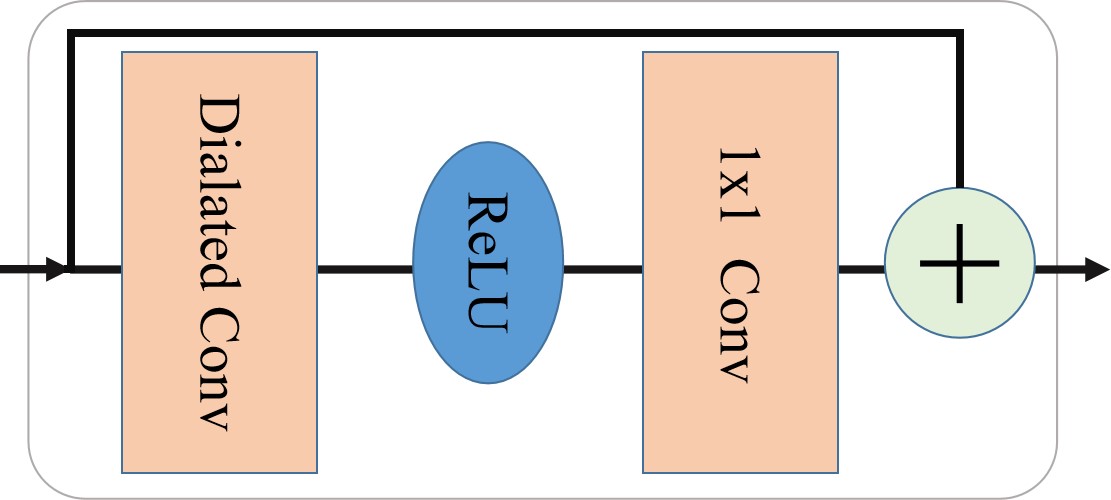}
  \caption{The architecture of dilated residual layer.}
  \label{fig:unit}
\end{figure}

The first part of TAN is a set of LCN unit for extracting each frame density maps, which will be described in Section \ref{sec:LCNN}. Formally, let $X=(x_1,...,x_T)$ be a video with $T$ frames. Each frame $x_i$ go through the LCN unit to produce the corresponding density map $f(x_i)$. In order to match the data dimension from the density map to the timing block group, we set the Reshape \& Concatenation unit. This unit transforms the density map $f(x_i)$ with size of ($M$,$N$) to into a one-dimensional vector $v_i$ with size of ($1$,$MN$).

The feature vectors are sent to a series of dilated residual blocks. The group of dilated residual block use the previous stage initial the next stage and use the next stage refines the previous stage. We define the frame orientation characteristics of the input video of the first stage as follows:
  \begin{equation}
  \begin{array}{l}
  {Y_0} = {x_{1:T}}\\
  {Y_s} = \mathcal F({Y_{s - 1}}),
  \end{array}
  \label{equ:fi}
  \end{equation}
where ${Y_s} $ is the output at $s$ stage and $\mathcal F$ is a dilated residual block. Each dilated residual block contains multiple dilated residual layers and the architecture of dilated residual layer is shown in Fig.~\ref{fig:unit}. The first is a dilated convolution with a receptive field, which helps in preventing the model from overfitting. Let $\mathbf{w}_{1,i}$ and ${b}_{1,i}$ be the filter weights and bias associated with the $i$-th dilated residual layer and $\mathbf{v}_i$ be the input, the output for location $l$ after the 1$D$ dilation is defined as
\begin{equation}
    \mathbf{\hat{v}}_i[l]=\sum_{\Delta l\in \mathcal{R}_d}\mathbf{w}_{1,i}[\Delta l]\cdot\mathbf{v}_i[l+\Delta l]+ {b}_{1,i},
\end{equation}
where $\mathcal{R}_d=\{-d,0,d\}$ construct the 1D filters with kernel size of 3 and $d=2^{i-1}$.

Then ReLU and $1 \times 1$ convolution are used to superimpose the weights and offset the output. The output of the whole dilated residual layer is
\begin{equation}
{\mathbf{{v}}_{i+1}} = {\mathbf{{v}}_{i}} + \mathbf{w}_{2,i}\cdot{\mathop{\rm ReLU}}{(\mathbf{\hat{v}}_i)} + {b}_{2,i},
\label{equ:Unit}
\end{equation}
where ${\mathbf{{v}}_{i+1}}$ is the output of layer $i$, $\mathbf{w}_{2,i}$ and ${b}_{2,i}$ are the weights and bias of the dilated convolution filters. The receptive field at the $i$-th dilated residual layer is $2^{i + 1} - 1$. A dilated residual block consists of three dilated residual layer, and we use this architecture to help provide more context to predict the result at each frame.

There are a few alternative choices to model the context with dilated convolution, such as dilated temporal convolution~\cite{lea2016temporal}, dilation with densely connection \cite{xu2018dense}, and dilated residual unit~\cite{farha2019ms}. In this paper, our design is based on the dilated residual unit for its computation efficiency. Our model aims to capture dependencies between current frame and the other video sequences, which helps smooth the prediction errors in the same video sequences.

To utilize the context information gain more effectively, we normalize the output of the last block and obtain a set of weight vectors. For the $t$-th video frame, the output of the last block is ${\mathbf{v}}(t) = [{v_{t - k}'^T}, \cdots ,v_t'^T \cdots ,{v_{t + k}'^T}]^T$ and the vector $v_i'$ represents the feature of $i$-th frame. We extract the weight from the normalization of the continuous frame features, \ie,
\begin{equation}
w_{\Delta t} =\frac{||v_{t+\Delta t}'||_1}{\sum\limits_{\Delta t=-k}^k||v_{t+\Delta t}'||_1}
\label{equ:Wj}
\end{equation}
We consider the original density map $f(x_t)$ again and the final density map $f'(x_t)$ is done by
\begin{equation}
f'(x_t) =\sum\limits_{\Delta t=-k}^k w_{\Delta t} f(x_{t+\Delta t})
\end{equation}
The final counting result for frame $t$ is computed by simply accumulating the density map $f'(x_t)$.

\vspace{0.08in}
\noindent\textbf{Loss function.} Learning of the parameters in each block is with two terms in the loss function, \ie,
\begin{equation}
{\mathcal L}_{block} = {{\mathcal L}_{mse}} + \lambda {{\mathcal L}_{SL1}},
\label{equ:Lme}
\end{equation}
where ${\lambda}$ is a model hyper-parameter to determine the contribution of the different terms.

The MSE loss is defined as
\begin{equation}
{{\mathcal L}_{mse}} = \frac{1}{N}\sum\limits_{i = 1}^N {{(C_{p} - C_{gt})^2}},
\label{equ:Lmse}
\end{equation}
where $N$ is the total amount of video frames, $C_p$ is the predicted counting value, and $C_{gt}$ is ground-truth.

While the MSE loss already performs well, we observe that the predictions for some of the videos contain a few over-segmentation errors. To further improve the quality of the predictions, we use an additional smoothing loss to reduce such over-segmentation issue. Here a Smooth-$L_1$ loss is employed:
\begin{equation}
{\mathcal L}_{SL1}(x,y) = \frac{1}{N}\left\{ {\begin{array}{*{20}{c}}
{\frac{1}{2}{{\left( {{x_i} - {y_i}} \right)}^2}}&{\mathrm{if}{\rm{ }}\left| {{x_i} - {y_i}} \right|{\rm{  < 1}}}\\
{\left| {{x_i} - {y_i}} \right| - \frac{1}{2}}&{\mathrm{otherwise}}
\end{array}} \right.
\label{equ:SL1}
\end{equation}

Several blocks will be applied in the TAN framework, and the loss function is the sum of ${\mathcal L}_{block}$ in each block.

\subsection{The LCN Unit}
\label{sec:LCNN}

The basic network in our framework is a convolutional neural network for crowd counting of a still image or a single video frame. In previous works, networks with multiple subnets and single branch are usually employed. Since we focus on video crowd counting problem in this paper, the inference speed is an important issue in real-world application and our goal is to use a small enough architecture to achieve a comparable result. Therefore, the single branch network with few parameters is preferred. We design a lightweight convolutional neural network (LCN) with 9 convolutional layers and 3 max pooling operations. The overall structure of LCN is illustrated in Table~\ref{table:LCN}. In our preliminarily experiments, we find that using more convolutional layers with small kernels is more efficient than using fewer layers with larger kernels for crowd counting, which is consistent with the observations from recent research on image recognition~\cite{Simonyan15}. Max pooling is applied for each $2\times2$ region, and Rectified linear unit (ReLU) is adopted as the activation function for its good performance. The network only consists of convolutional blocks with $3\times3$ kernel and max-pooling layers instead of more sophisticated architecture, which aims to accelerate computational speed. We also limit the number of filters on each layer to reduce the computational complexity. Finally, we adopt filters with a size of 1 ${\times}$ 1 to generate the features vector. The end-to-end architecture makes the training procedure easier. The loss function of LCN is defined as
\begin{equation}
{\mathcal L}_{LCN} = \frac{1}{{2N}}\sum\limits_{i = 1}^N {||f({x_i}) - {F({x_i})}||_2^2},
\label{equ:loss}
\end{equation}
where $N$ is the number of training images, and $F({x_i})$ is the ground truth density map of image $x_i$, and ${f({x_i})}$ is the estimated density map for $x_i$. The overall network parameter size of LCN is less than $5 \times 10^4$, and the network can obtain real-time speed under a CPU environment. As will be shown in the experiments, within a small size of parameters, our model can still achieve a competitive result compared with previous approaches.

\begin{table}[t]
\caption{Configuration of LCN.}
\label{table:LCN}
\centering
\begin{tabular}{@{}|l|c|c|c|c|@{}}
\hline
Layer & Kernel size & Channel & Dilation rate & Output \\ \hline
conv1 & $3\times3$         & 8                 & 1             & 1      \\ \hline
max-pool1 & $2\times2$ & -                 & -             & 1/2    \\ \hline
conv2 & $3\times3$         & 16                & 1             & 1/2    \\ \hline
conv3 & $3\times3$         & 16                & 1             & 1/2    \\ \hline
max-pool2 & $2\times2$ & -                 & -             & 1/4    \\ \hline
conv4 & $3\times3$         & 32                & 1             & 1/4    \\ \hline
conv5 & $3\times3$         & 32                & 1             & 1/4    \\ \hline
conv6 & $3\times3$         & 32                & 1             & 1/4    \\ \hline
max-pool3 & $2\times2$ & -                 & -             & 1/8    \\ \hline
conv7 & $3\times3$         & 16                & 1             & 1/8    \\ \hline
conv8 & $3\times3$         & 8                 & 1             & 1/8    \\ \hline
conv9 & $1\times1$         & 1                 & 1             & 1/8    \\ \hline
\end{tabular}
\end{table}

\subsection{Implementation Details}
\label{sec:ID}

\begin{figure}
  \centering
  \includegraphics[width=.8\linewidth]{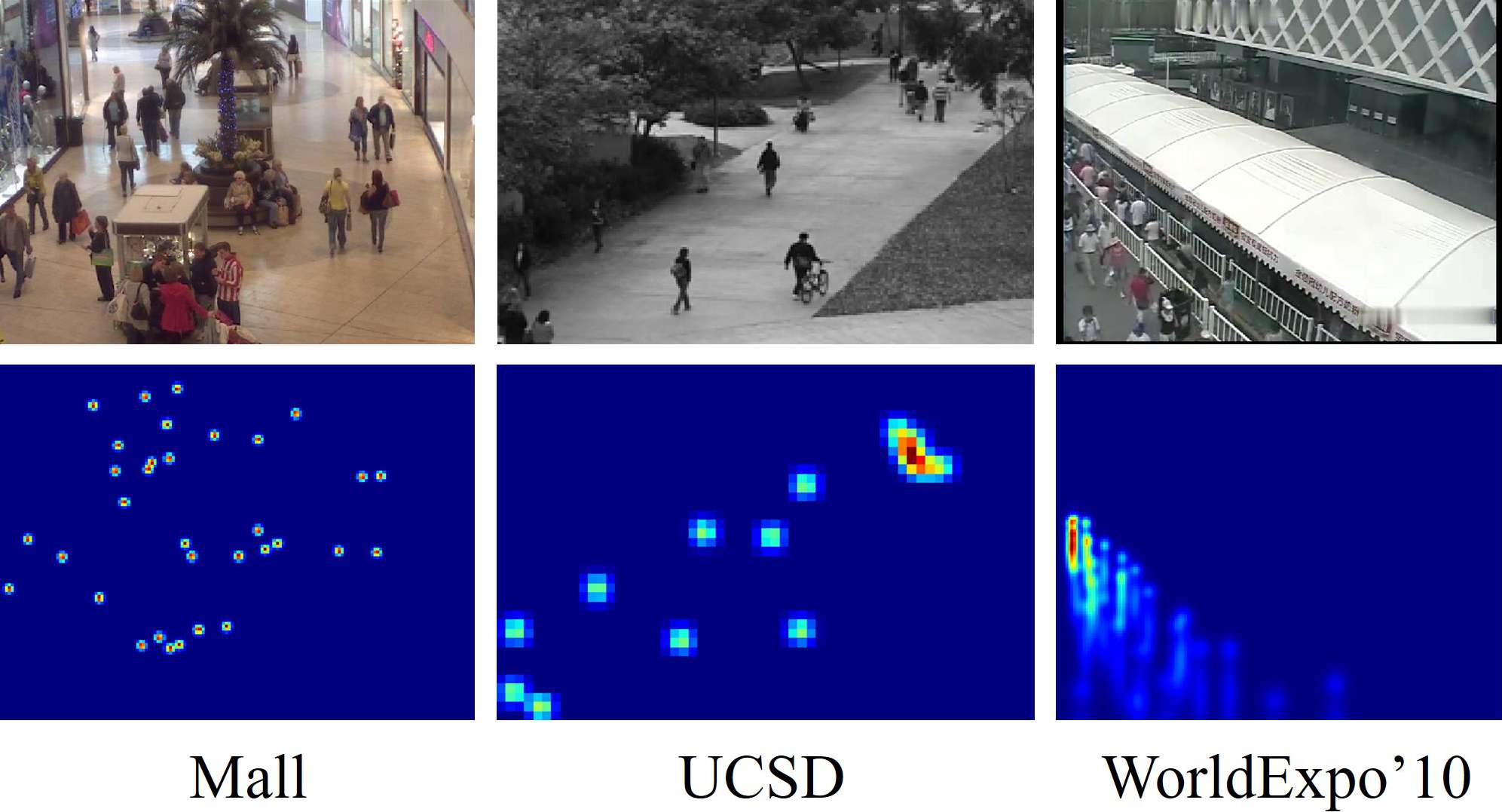}
  \caption{Ground-truth density map for different datasets.}
  \label{fig:DM}
\end{figure}

\noindent\textbf{Ground truth generation.} There is significate diversity among different crowd counting datasets and thus we use the geometry-adaptive kernels to generate density maps from the ground truth. The geometry-adaptive kernels are defined as
\begin{equation}
F(x) = \sum\limits_{i = 1}^{N_t} {\delta (x - {o_i})}  \times {G_{\sigma_i}}(x).
\label{equ:F}
\end{equation}
Given object $o_i$ in the target set $\{o_1,o_2,...,o_{N_t}\}$, we calculate $k$ nearest neighbors to determine $d_i$. For the pixel position $i$ in the image, we use a Gaussian kernel with a parameter of ${\sigma _i} = \beta {\bar d_i}$ to generate the density map $F(x)$.

In our experiments, we create density maps with the fixed kernel of 17 for UCSD dataset and 15 for others. We also follow the previous work~\cite{Zhang2015Cross} to create density maps using Region of Interest (ROI) and the perspective map to deal with the WorldExpo'10 dataset.

\vspace{0.08in}
\noindent\textbf{Data augmentation.}
We consider data augmentation based on the original information of the data. For the training of LCN, the insufficient number of training samples is one important issue. Thus, we follow the data enhancement method in \cite{Li2018CSRNet} to deal with image data. Nine color patches are cut from each image in different positions and the size is $\frac{1}{4}$ of the original image. The first four tiles contain three-quarters of the images without overlapping, while the other five tiles are randomly cropped from the input image. After that, we mirrored the patches to double the training set. We do not apply any data enhancement for the video dataset, as we would like to consider more context information of the video frames by using our model.

\vspace{0.08in}
\noindent\textbf{Training details.}
Our temporal model is implemented using PyTorch. To train the LCN, we first initialize the layers of the network using a Gaussian distribution from standard deviation of 0.01. We then set the learning rate of $10^{-5}$ for all the datasets as initial, and use Adam~\cite{kingma2014adam} for training. For the training of TAN, we use Adam optimizer with the learning rate of 0.0005.

\section{Experiments}
\label{sec:exp}

\begin{table}
\centering
\caption{Statistics of the datasets. Num: the number of video frames or images; Avg: the average labeled pedestrian number in the dataset; Total: the total labeled pedestrians number.}
\label{table:Dataset}
\small{
\begin{tabular}{@{}lccccc@{}}
\toprule
Dataset   &Type & Resolution  & Num   & Avg & Total  \\ \midrule
UCSD      &Video & 158$\times$238    & 2,000    & 24.9    & 49,885  \\
Mall      &Video & 640$\times$480     & 2,000    & 31.2    & 62,315  \\
WorldExpo &Video & 576$\times$720     & 3,980   & 50.2    & 199,923 \\\hline
ShanghaiTech A &Image & Varied     & 482     & 501    & 241,677 \\
ShanghaiTech B &Image & 768$\times$1024     & 716    & 123    & 88,488 \\
UCF\_CC\_50 &Image & Varied     & 50     & 1279    & 63,974 \\
\bottomrule
\end{tabular}
}
\end{table}

\begin{table}
\centering
\caption{Crowd counting results on Mall and UCSD. The performance of \cite{chan2008privacy,chen2012feature,chen2013cumulative} are from \cite{xiong2017spatiotemporal}.}
\label{table: MALL}
\small{
\begin{tabular}{@{}p{130px}cccc@{}}
\toprule
Method & \multicolumn{2}{c}{MALL} & \multicolumn{2}{c}{UCSD}  \\\cline{2-5}
                   & MAE          & MSE         & MAE          & MSE            \\ \cline{1-5}
Gaussian process regression~\cite{chan2008privacy}            &3.72    &20.1 & 2.24   & 7.97\\
Ridge regression~\cite{chen2012feature}                        &3.59    &19.0 & 2.25   & 7.82\\
\small{Cumulative attribute regression~\cite{chen2013cumulative}}    &3.43    &17.7 & 2.07   & 6.90\\\hline
ConvLSTM~\cite{xiong2017spatiotemporal}                     &2.24    &8.5 & 1.30 & 1.79\\
\small{Bidirectional ConvLSTM~\cite{xiong2017spatiotemporal} }      &2.10   & 7.6 & 1.13 & 1.43\\
ST-CNN~\cite{miao2019st}     &4.03   & 5.87 & - & -\\ \hline
TAN                                                        & {2.03}   &  {2.6}  &  {1.08} &  {1.41}\\ \bottomrule
\end{tabular}
}
\end{table}

We evaluate the Temporal Aware Network on three video crowd counting benchmarks, \ie, Mall~\cite{chen2012feature}, UCSD~\cite{chan2008privacy}, and WorldExpo'10~\cite{Zhang2015Cross}. Fig. \ref{fig:DM} illustrates their typical scenes. To examine the efficiency of the basic network LCN, we also conduct the image-level analysis on ShanghaiTech~\cite{Zhang_2016_CVPR} and UCF\_CC\_50~\cite{Idrees2013Multi} datasets, since there are no time-related information. Basic statistics of the datasets are summarized in Table~\ref{table:Dataset}.

Following existing state-of-the-art methods, we use the mean absolute error (MAE) and mean squared error (MSE) to evaluate the performance, which are defined as
\begin{equation}
MAE = \frac{1}{N}\sum\limits_{i = 1}^N {\left| {{C_i} - C_i^{GT}} \right|},
\label{equ:mae}
\end{equation}
\begin{equation}
MSE = \sqrt {\frac{1}{N}\sum\limits_{i = 1}^N {{{\left| {{C_i} - C_i^{GT}} \right|}^2}}}.
\label{equ:mse}
\end{equation}
Here $N$ is the number of testing images, $C_i$ and $C_i^{GT}$ are the estimated people count and ground truth people count in the $i$-th image respectively.

There are a few hyper-parameters in TAN, such as the number of video frames for temporal modeling and dilated residual blocks. In this section, we use 5 video frames for the temporal modeling and 3 blocks as the default setting. The effect of these settings will be evaluated thoroughly in the ablation study. We also report the number of neural networks parameters (Params) for comparison.

\begin{figure*}
  \centering
  \includegraphics[height=1.2in,width=.473\linewidth]{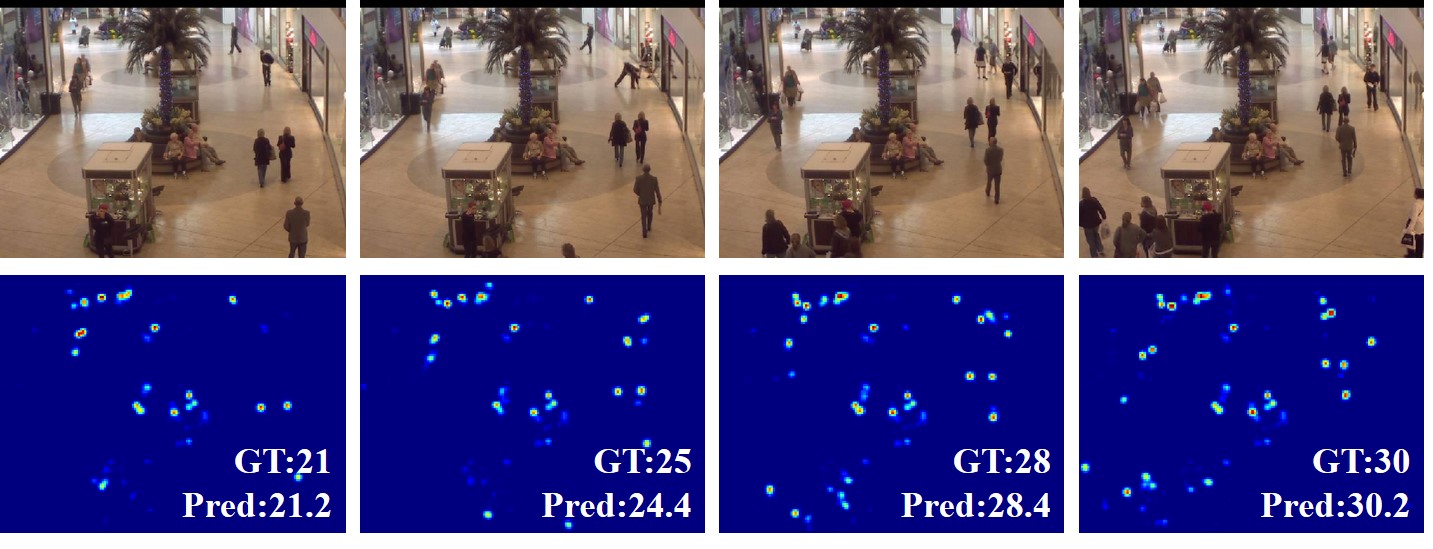}  \hspace{0.05in}
  \includegraphics[height=1.2in,width=.473\linewidth]{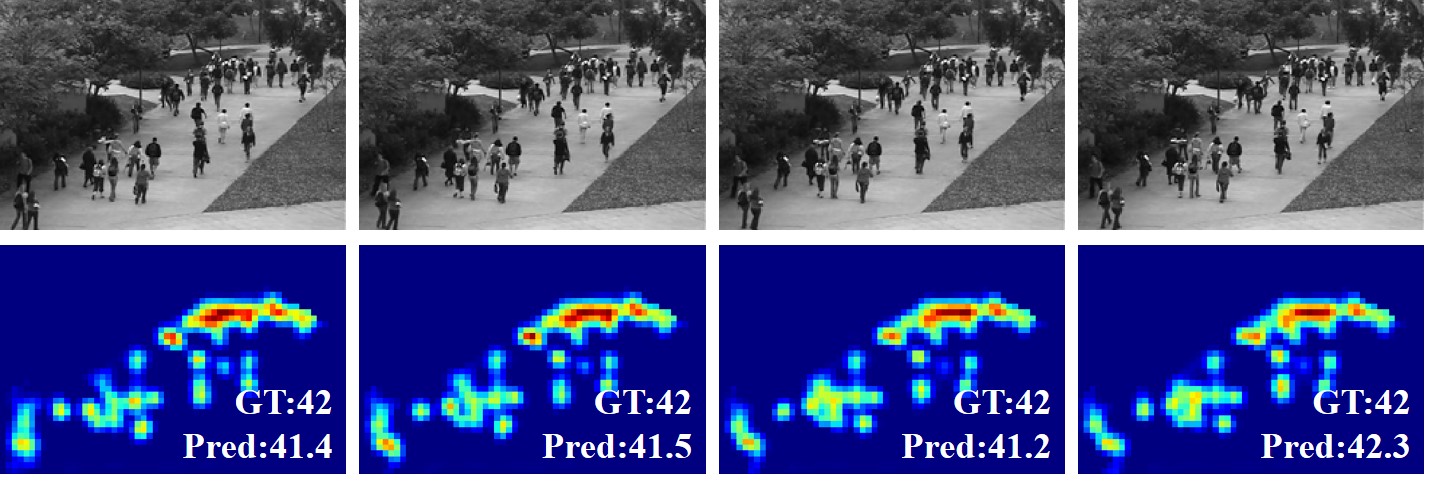}\\
  \small{(a) MALL}\hspace{3.1in}\small{(b) UCSD}\\
  \vspace{0.06in}
  \includegraphics[height=1.2in,width=.475\linewidth]{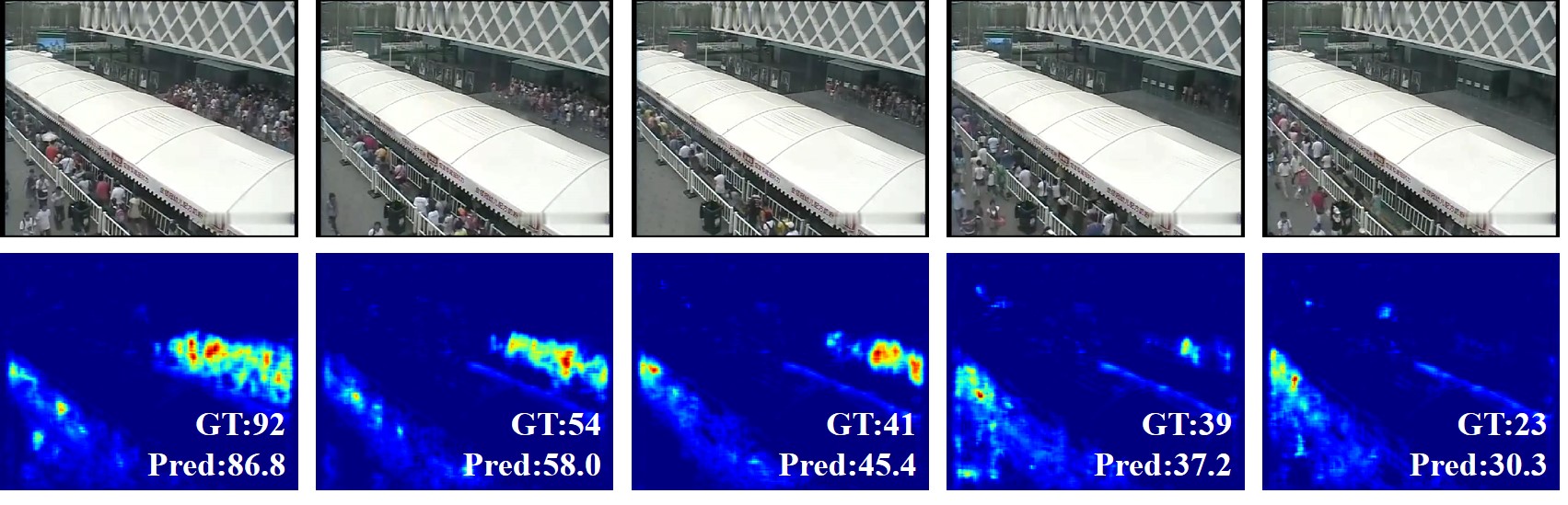}  \hspace{0.02in}
  \includegraphics[height=1.2in,width=.475\linewidth]{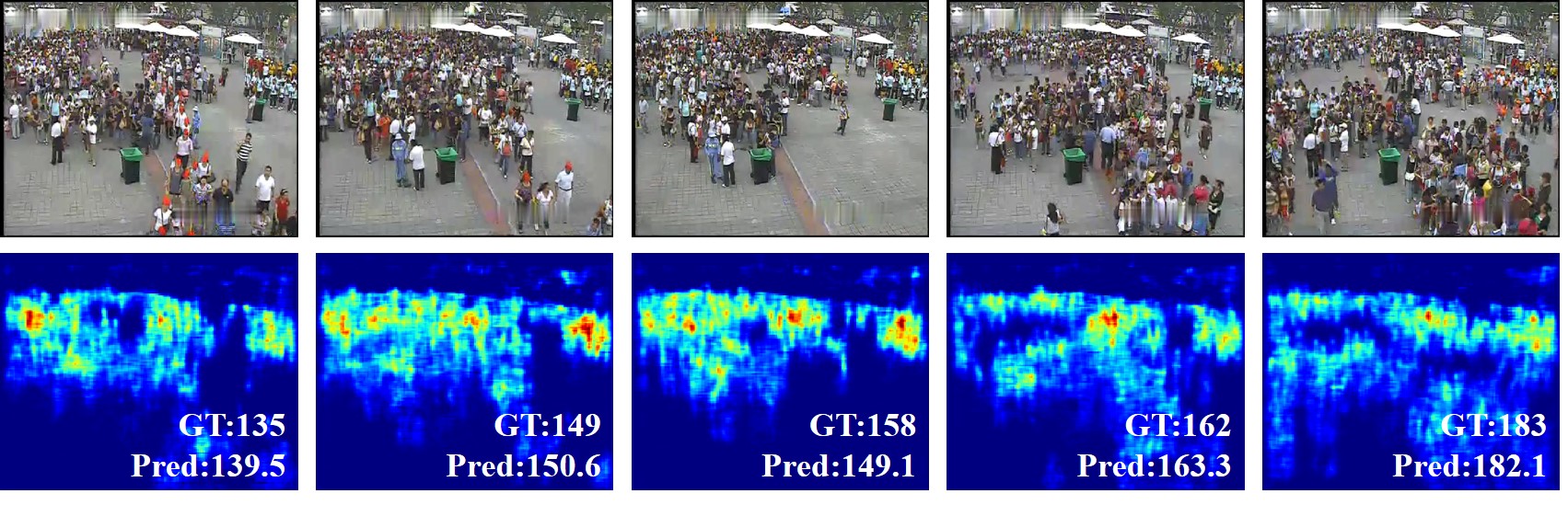}\\
  \small{(c) WorldExpo'10 (Scene 3, Scene 4)}
  \caption{Qualitative results on the sample snippets of the video datasets.}
  \label{fig:video}
\end{figure*}

\subsection{Mall Dataset}

We first report the results on the Mall dataset summarized in Table \ref{table: MALL}-Left. The experiments follow the same setting as~\cite{chen2012feature}, in which the first 800 frames are used for training and the remaining 1,200 frames are used for testing. we compare the TAN with the methods of using spatialtemporal information as well, including the regression-based methods~\cite{chan2008privacy,chen2012feature,chen2013cumulative} and the temporal-based methods~\cite{xiong2017spatiotemporal,miao2019st}. As shown in the table, using the proposed TAN leads to the MAE of 2.03 and MSE of 2.6, which are significantly higher than the baseline approaches. We demonstrate some predicted density maps as well as their corresponding counting results with TAN in Fig. \ref{fig:video}(a).

\begin{table}
\centering
\caption{The MAE of different scenes on the WorldExpo'10 dataset.}
\label{table:maeWD}
\small{
\begin{tabular}{@{}p{66px}p{12px}<{\centering}p{12px}<{\centering}p{12px}<{\centering}p{12px}<{\centering}p{12px}<{\centering}p{12px}<{\centering}p{45px}<{\centering}@{}}
\toprule
Method                                                     & S1  & S2   & S3   & S4   & S5  & Avg. &  {Params}\\ \midrule
ic-CNN ~\cite{babu2018divide}                              & 17.0& 12.3  & 9.2   &  {8.1}  & 4.7  & 10.3  & {$16.82 \times10^6$}\\
\small{D-ConvNet~\cite{Shi_2018_CVPR}}                     &  {1.9} & 12.1  & 20.7  & 8.3  & 2.6  & 9.1   & {$16.26 \times 10^6$}\\
CSRNet~\cite{Li2018CSRNet}                                 & 2.9 &  {11.5}  &  {8.6}   & 16.6 & 3.4  & 8.6   & {$16.26 \times10^6$}\\
ACSCP~\cite{Shen_2018_CVPR}                                & 2.8 & 14.1  & 9.6   & 8.1  &  {2.9}  &  {7.5}   & {$5.10 \times 10^6$}\\
\hline
ConvLSTM~\cite{xiong2017spatiotemporal}                    & 7.1 & 15.2  & 15.2  & 13.9 & 3.5  & 10.9  &-\\
{\footnotesize{Bi-ConvLSTM}}~\cite{xiong2017spatiotemporal}       & 6.8 &  {14.5}  & 14.9  & 13.5 &  {3.1}  & 10.6  &-\\
ST-CNN~\cite{miao2019st}       & 5.2 & 16.5  & 9.9  & 8.4 & 6.2  & 9.3  &-\\
\hline
TAN                                                        &  {2.8} & 18.1  &  {9.6}   &  {7.5}  & 3.6  &  {8.3} & {$0.047 \times 10^6$}\\ \bottomrule
\end{tabular}
}
\end{table}

\subsection{UCSD Dataset}

Following the convention of the existing works~\cite{chan2008privacy}, we use frames 601-1400 of the UCSD dataset as the training data and the remaining 1200 frames as the test data. As the region of interest (ROI) and perspective map are provided, the intensities of pixels out of ROI is set to zero, and we also use ROI to supervise the last convolution layer. Results on the UCSD dataset are presented in Table \ref{table: MALL}-Right. Same as the experiments on the Mall dataset, TAN shows better results than the LSTM-based crowd counting approaches. Some counting results with TAN on sample snippets are shown in Fig. \ref{fig:video}(b).

\subsection{WorldExpo'10 Dataset}

The WorldExpo'10 dataset consists of 3,980 annotated frames from 1,132 video sequences captured by 108 different surveillance cameras during the 2010 Shanghai World Expo. The training set includes of 3,380 annotated frames from 103 scenes, while the testing images are extracted from other five different scenes with 120 frames per scene. Following the settings of \cite{Zhang2015Cross}, MAE is used as the evaluation metric. Table~\ref{table:maeWD} lists the per-scene performance of TAN and previous approaches. Among these approaches, the first group is the state-of-the-art methods with pre-trained models~\cite{babu2018divide,Shi_2018_CVPR,Li2018CSRNet}  or more complex network designs~\cite{Shen_2018_CVPR}. Our results are comparable with these approaches for four scenes (except in Scene 2), while the parameter size of the TAN is order-of-magnitude smaller than all of these methods. And the results of TAN is also better than that of the temporal-based methods~\cite{xiong2017spatiotemporal,miao2019st}. The qualitative results on one of the testing scenes are illustrated in Fig. \ref{fig:video}(c).

\begin{table}[t]
\centering
\caption{The metrics of the LCN comparing with the previous approaches on ShanghaiTech Part A \& Part B (SHA \& SHB) and UCF\_CC\_50 (UCF).}
\label{table:maeABUCF}
\small{
\begin{tabular}{@{}lccccc@{}}
\toprule
Method                   &     & SHA & SHB & UCF   &  {Params}              \\ \midrule
\multirow{2}{*}{ic-CNN ~\cite{babu2018divide}}  & MAE & 68.5           & 10.7           & 260.9 & \multirow{2}{*}{ {$16.82 \times 10^6$}} \\
                         & MSE & 116.2          & 16.0           & 365.5 &                        \\ 
\multirow{2}{*}{D-ConvNet~\cite{Shi_2018_CVPR}}   & MAE & 73.5           & 18.7           & 288.4 & \multirow{2}{*}{ {$16.26 \times 10^6$}} \\
                         & MSE & 112.3          & 26.0           & 404.7 &                        \\ 
\multirow{2}{*}{CSRNet~\cite{Li2018CSRNet}}     & MAE & 68.2           & 10.6           & 266.1 & \multirow{2}{*}{ {$16.26 \times 10^6$}} \\
                         & MSE & 115.0          & 16.0           & 397.5 &                        \\ 
\multirow{2}{*}{ACSCP~\cite{Shen_2018_CVPR}}   & MAE & 75.7           & 17.2           & 291.0 & \multirow{2}{*}{ {$5.10 \times 10^6$}}  \\
                         & MSE & 102.7          & 27.4           & 404.6 &                        \\ 
\hline
\multirow{2}{*}{BSAD~\cite{Huang2017Body}}    & MAE & -              & 20.2           & 409.5 & \multirow{2}{*}{ {$1.30 \times 10^6$}}  \\
                         & MSE & -              & 35.6           & 563.7 &                        \\ 
\multirow{2}{*}{TDF-CNN~\cite{babu2018top}} & MAE & 97.5           & 20.7           & 354.7 & \multirow{2}{*}{ {$1.15 \times 10^6$}}  \\
                         & MSE & 145.1          & 32.8           & 491.4 &                        \\ 
\multirow{2}{*}{MCNN~\cite{Zhang_2016_CVPR}}    & MAE & 110.2          & 26.4           & 377.6 & \multirow{2}{*}{ {$0.13 \times 10^6$}}  \\
                         & MSE & 173.2          & 41.3           & 509.1 &                        \\ \hline
\multirow{2}{*}{LCN}    & MAE & 93.3           & 15.1           & 262.0 & \multirow{2}{*}{ {$0.032 \times 10^6$}} \\
                         & MSE & 157.0          & 23.3           & 358.6 &                        \\ \bottomrule
\end{tabular}
}
\end{table}

\begin{figure}[t]
  \centering
  \includegraphics[width=3in]{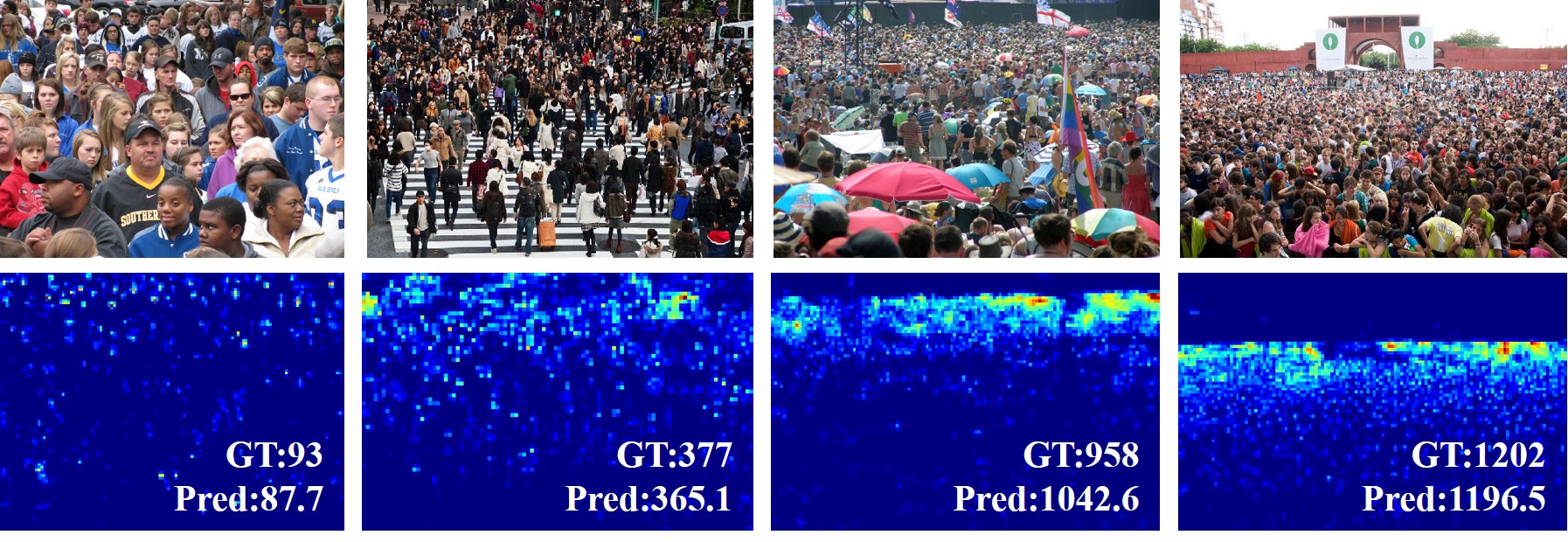}\\
  \small{(a) ShanghaiTech Part A}\\
  \vspace{0.06in}
  \includegraphics[width=3in]{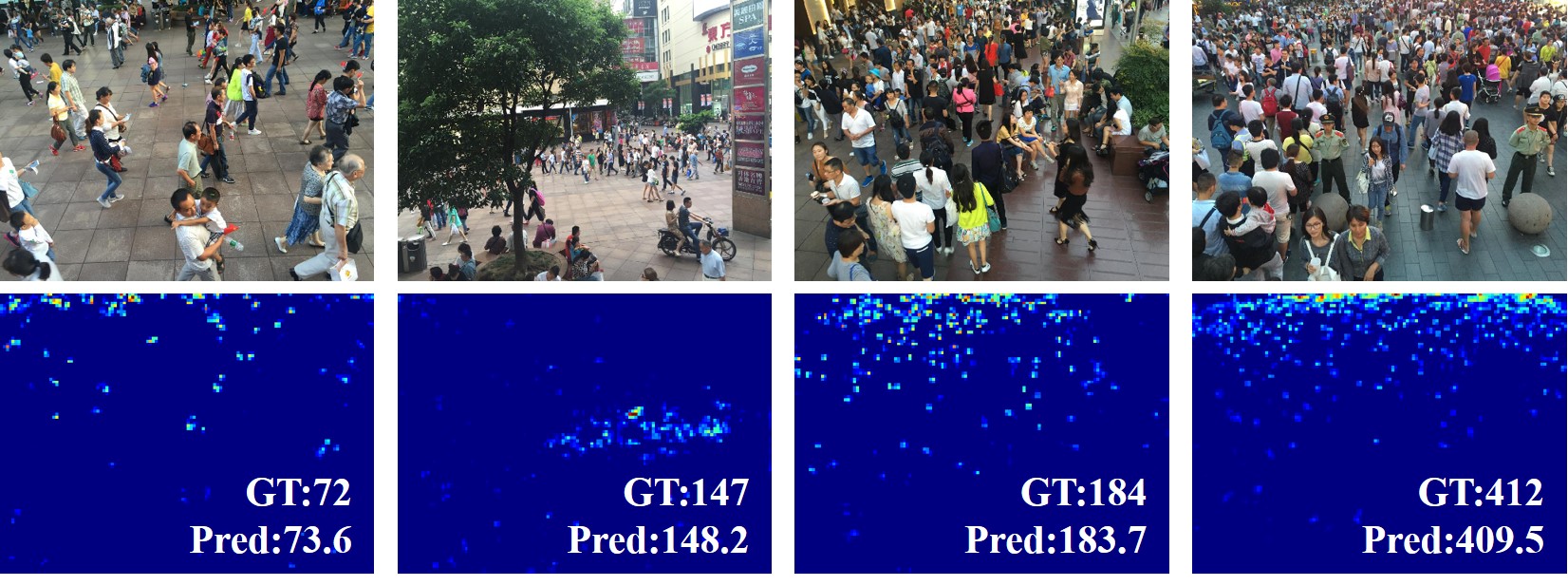}\\
  \small{(b) ShanghaiTech Part B}\\
  \vspace{0.06in}
  \includegraphics[width=3in]{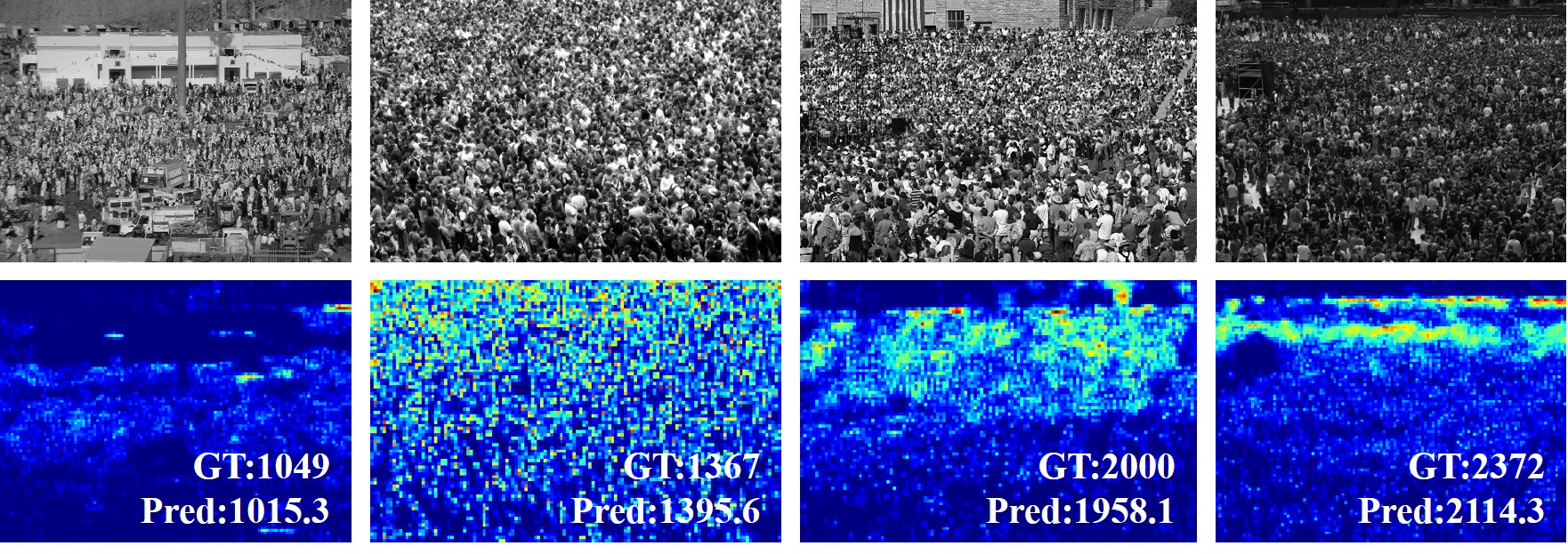}\\
  \small{(c) UCF\_CC\_50}
  \caption{Density maps and predicted counting by the basic network.}
  \label{fig:still}
\end{figure}

\begin{table}[t]
\centering
\caption{ {Evaluation of counting results w.r.t. the number of video frames.}}
\label{table:ablationstudy1}
\footnotesize{
\begin{tabular}{p{0.98in}p{0.38in}<{\centering}p{0.38in}<{\centering}p{0.38in}<{\centering}}
\toprule
Method                          & Dataset & MAE  & MSE  \\ \hline
\multirow{2}{*}{LCN}     & UCSD    & 1.13 & 1.47 \\
                                & MALL    & 2.23 & 2.85 \\\hline
\multirow{2}{*}{TAN (3 frames)} & UCSD    & 1.10 & 1.43 \\
                                & MALL    & 2.06 & 2.62 \\\hline
\multirow{2}{*}{TAN (5 frames)} & UCSD    & 1.08 & 1.41 \\
                                & MALL    & 2.03 & 2.57 \\\hline
\multirow{2}{*}{TAN (7 frames)} & UCSD    & 1.08 & 1.41 \\
                                & MALL    & 2.03 & 2.59 \\\hline
\multirow{2}{*}{LCN (5 frames average)} & UCSD    & 1.12 & 1.45 \\
                                & MALL    & 2.20 & 2.60 \\\bottomrule
\end{tabular}
}
\end{table}

\begin{table}[t]
\centering
\caption{ {Evaluation of counting results w.r.t. the block number.}}
\label{table:ablationstudy2}
\footnotesize{
\begin{tabular}{p{0.98in}p{0.38in}<{\centering}p{0.38in}<{\centering}p{0.38in}<{\centering}}
\toprule
Method                          & Dataset & MAE  & MSE  \\ \hline
\multirow{2}{*}{TAN (1 block)}     & UCSD    & 1.09 & 1.43 \\
                                & MALL    & 2.03 & 2.58 \\\hline
\multirow{2}{*}{TAN (3 blocks)} & UCSD    & 1.08 & 1.41 \\
                                & MALL    & 2.03 & 2.57 \\\hline
\multirow{2}{*}{TAN (5 blocks)} & UCSD    & 1.08 & 1.41 \\
                                & MALL    & 2.08 & 2.65 \\\bottomrule
\end{tabular}
}
\end{table}

\begin{table}[t]
\centering
\caption{Comparison of different temporal modeling approaches.}
\label{table:temporal}
\footnotesize{
\begin{tabular}{p{0.98in}p{0.38in}<{\centering}p{0.38in}<{\centering}p{0.38in}<{\centering}}
\toprule
Method                          & Dataset & MAE  & MSE  \\ \hline
\multirow{2}{*}{LCN + LSTM}     & UCSD    & 1.21 & 1.69 \\
                                & MALL    & 2.23 & 3.80 \\\hline
\multirow{2}{*}{LCN + BI-LSTM} & UCSD    & 1.11 & 1.48 \\
                                & MALL    & 2.09 & 3.07 \\\hline
\multirow{2}{*}{TAN}        & UCSD    & 1.08 & 1.41 \\
                                & MALL    & 2.03 & 2.60 \\ \bottomrule
\end{tabular}
}
\end{table}

\subsection{Ablation Study}
In this section, we evaluate some parameters and alternative implementations of the proposed framework.

\vspace{0.08in}
\noindent\textbf{LCN.} We first evaluate the performance of LCN and compare it with several approaches. As most of the previous approaches report results on ShanghaiTech and UCF\_CC\_50, here we also conduct the comparison on these datasets and Table \ref{table:maeABUCF} reports the metrics. Among these approaches, the first group are also the state-of-the-art methods with more complex networks~\cite{babu2018divide,Shi_2018_CVPR,Li2018CSRNet,Shen_2018_CVPR}. Our results are comparable with these approaches, while our model size is much more compact. The second group contains several networks with compact structure, including MCNN~\cite{Zhang_2016_CVPR}, BSAD~\cite{Huang2017Body}, and TDF-CNN~\cite{babu2018top}. From the table it is clear that LCN outperforms all these approaches. Fig. \ref{fig:still} illustrates some examples using LCN on both datasets, including crowd images, predicted density maps, and the counting results.

\vspace{0.08in}
\noindent\textbf{Number of video frames for temporal modeling.} As shown in Table \ref{table:ablationstudy1}, we compare the performance of our framework with a varying number of video frames for the temporal modeling. We observe performance gains when the number of considered video frames increases from three to five. Using more frames does not improve performance since the number of crowds varies along the time pass. Another intuitive way to add the temporal information is to smooth the density maps or counting numbers of neighboring frames. However, as shown in the table, this strategy improves the performance of the single frame model, but is not as good as the proposed TAN approach.

\vspace{0.08in}
\noindent\textbf{Number of dilated residual blocks.} We also evaluate the effect of dilated residual block numbers in the TAN model. As shown in Table \ref{table:ablationstudy2}, the best trade-off is obtained by using three dilated residual blocks. Compared to using a single block, more blocks can boost performance. However, when the number gets larger, in some case the performances are decreased. This is probably because complex neural networks lead to underfitting when the scale of training data is limited.

\vspace{0.08in}
\noindent\textbf{Temporal modeling.} We compare our temporal aware network with previous LSTM based approaches by incorporating LCN with them. As shown in Table~\ref{table:temporal}, the results of TAN are better than LCN with LSTM or Bi-directional LSTM. This also proves that our temporal modeling can capture temporal relations better than LSTM.

\vspace{0.08in}
\noindent\textbf{Timing.} Recall that our goal is to build a compact model for fast crowd counting in the videos based on the proposed lightweight network. The parameter size of LCN and TAN are 0.032M and 0.047M respectively. For a video with the resolution of  $320\times240$ pixels, the TAN model achieves 120 FPS detection speed on an Nvidia TITAN X GPU and during inference it only consumes less than 500MB GPU memory. Our approach can provide real-time (25FPS) crowd counting speed with a moderate Intel Core i5-8400 desktop CPU.

\section{Conclusions}
\label{sec:conclusion}

We proposed the Temporal Aware Network with the LCN unit toward fast video crowd counting. The novel lightweight architecture is able to produce good performance with the compact network. We showed that by leveraging contexture information of the video contents, promising results are achieved for video crowd counting benchmark. We also achieved the real-time inference on a moderate commercial CPU by 25 FPS.

\bibliographystyle{IEEEtran}
\bibliography{total}

\end{document}